\pdfoutput=1

\documentclass[letterpaper, 10 pt, conference]{ieeeconf}  

\IEEEoverridecommandlockouts                              

\usepackage{graphicx, graphics} 
\usepackage{times} 
\usepackage{amsmath} 
\usepackage{amssymb}  
\usepackage{color}
\usepackage{mathrsfs}

\usepackage{url}
\usepackage{booktabs}

\definecolor{mygray}{gray}{0.6}

\usepackage{hscc2017}

\usepackage[labelformat=simple]{subfig}

\title{\LARGE \bf
Dynamic Walking on Slippery Surfaces:\\
{\normalsize Demonstrating Stable Bipedal Gaits with Planned Ground Slippage$^*$}
}

\author{Wen-Loong Ma$^{1}$, Yizhar Or$^{2}$ and Aaron D. Ames$^{3}$
\thanks{$^{1}$W. Ma is with Mechanical Engineering, California Institute of Technology, Pasadena, CA, USA.
   	{\tt\small wma@caltech.edu}}%
\thanks{$^{2}$Y. Or is with the faculty of Mechanical Engineering, Technion - Israel Institute of Technology, Haifa, Israel. 	
   	{\tt\small izi@technion.ac.il}}%
\thanks{$^{3}$A. Ames is with the faculty of Mechanical Engineering and Control + Dynamical Systems, California Institute of Technology, Pasadena, CA, USA.
   	{\tt\small aames@caltech.edu}}
\thanks{$*$This work is supported by NSF grant 1724464, 1544332, 1724457, and Disney Research LA. The work has been conducted while Y. Or was hosted by A. D. Ames and AMBER lab at Caltech during his sabbatical leave from the Technion.}
}

\begin{document}
\maketitle
\thispagestyle{empty}
\pagestyle{empty}
{\abstract Dynamic bipedal robot locomotion has achieved remarkable success due in part to recent advances in trajectory generation and nonlinear control for stabilization. A key assumption utilized in both theory and experiments is that the robot's stance foot always makes no-slip contact with the ground, including at impacts. This assumption breaks down on slippery low-friction surfaces, as commonly encountered in outdoor terrains, leading to failure and loss of stability. In this work, we extend the theoretical analysis and trajectory optimization to account for stick-slip transitions at point foot contact using Coulomb's friction law. Using AMBER-3M planar biped robot as an experimental platform, we demonstrate for the first time a slippery walking gait which can be stabilized successfully both on a lubricated surface and on a rough no-slip surface. We also study the influence of foot slippage on reducing the mechanical cost of transport, and compare energy efficiency in both numerical simulations and experimental measurements.
}	

\section{INTRODUCTION}
Tremendous progress in realizing robust bipedal robot locomotion has been achieved in the last decade. This is in part due to successful combination of theoretical modeling and analysis using the framework of hybrid systems \cite{ames-hi, Grizzle20103D}, application of advanced methods of nonlinear control \cite{Khalil2002Nonlinear,Westervelt2007a}, as well as careful mechanical design and hardware implementation on various experimental platforms such as AMBER-3M \cite{ambrose2017toward}, DURUS \cite{reher2016algorithmic} and Cassie \cite{Da16from}. Underlying all of these results, along with successes for robots using other paradigms such as ZMP \cite{Vukobratovic2004ZMP, tedrakeZMP} and spring-loaded inverted pendulum (SLIP) based models \cite{PoulakakisFormal, VejdaniSLIP}, is the assumption that the foot does not slip. Thus, in all of these cases, the foot acts as a stationary pivot point. 
%
While this assumption may easily hold in sterile laboratory environments where the floors can be chosen with sufficiently high friction, it becomes impractical on natural outdoor terrains, wherein there are a plethora of slippery or slightly granulated irregular surfaces. 
Success in challenging the stationary contact point assumption include multi-contact walking \cite{zhaomulti} and bipedal running \cite{ma2017bipedal, Sreenath2013}.

\begin{figure}[t]
\vspace{2mm}
	\begin{center}
		\includegraphics[width=0.4\textwidth]{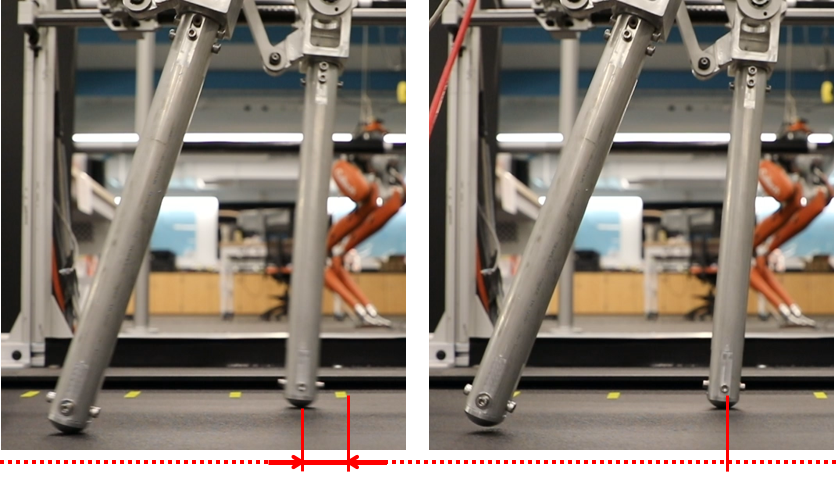}
		\caption{{\small Slippage in the beginning of a step: pre-slip on the left and post-slip on the right.}}
		\label{fig:robot}
	\end{center}
	\vspace{-8mm}
\end{figure}

The goal of this paper is to address this fundamental assumption of no slippage by embracing its violation while still being able to demonstrate the ability to achieve stable walking experimentally. In legged robots, foot slippage is often treated as an external disturbance which should be avoided at the gait planning stage \cite{chang1993sliding,kajita2004biped}, or detected and recovered in real-time by feedback control at the experimental implementation stage \cite{kaneko2005slip,vazquez2013experimental}. Some of the most famous examples are Boston Dynamics' robots BigDog \cite{bigdog} and SpotMini \cite{spot} successfully recovering from slippage. Conversely, legged animals across a wide range of scales show impressive adaptability to slippery surface on natural terrains. Stick insects confronted with a slippery surface modulate their motor outputs to produce normal walking gaits, despite a drastic change in the loads that these limbs experience \cite{gruhn2009straight}. Slippage in bipedal running of Guinea fowl has been studied in \cite{clark2011slipping}, showing that falling on slippery surfaces is a strong function of both speed and limb posture at touchdown. Several works in human biomechanics literature study the conditions that cause slipping \cite{moyer2006gait}, its consequences \cite{tinetti1997falls} and dynamics \cite{strandberg1981dynamics}. Finally, \cite{spencehoofSICB2007} has measured feet motion in galloping gaits of horses on outdoor racing terrains and found significant phase of hoof slippage.

Recent theoretical work has incorporated slippage into classic simple planar models of legged locomotion both in passive dynamic and actuated walking --- the rimless wheel \cite{gamus2015dynamic}, compass biped \cite{gamus2015dynamic,gamus2013analysis} and SLIP \cite{or2016analysis}. The models use Coulomb's friction law and account for stick-slip transitions and friction-bounded inelastic impacts, which add complexity to the system's multi-domain hybrid dynamics. 
Investigating the influence of friction on both passive dynamics down a slope and open-loop actuated walking, it has been found in \cite{gamus2015dynamic,gamus2013analysis} that upon decreasing the friction coefficient, periodic solutions with stick-slip transitions begin to evolve while their orbital stability decreases until reaching stability loss for too low friction. Nonetheless, stability can be recovered when adding simple PD control to track a reference trajectory. In addition, it has been found in \cite{gamus2015dynamic,gamus2013analysis,or2016analysis} that periodic solutions with slipping impact showed a significant reduction in energetic cost of transport compared to their no-slip counterparts. Nonetheless, these promising theoretical results have never been tested and implemented experimentally on legged robots. 

In this work, we bridge this gap by presenting, for the first time, an experimental realization of stable planar bipedal robotic walking on a slippery surface. First, we extend the formulation of the planar bipedal walking as a hybrid system in order to account for slippage and Coulomb friction inequalities, leading to multi-domain system with stick-slip transitions and stick/slip impact laws. Then we utilize the nonlinear programming (NLP) toolbox FROST developed in \cite{hereid2018dynamic} in order to generate gaits with slipping impact that are amenable to feedback control using the concept of hybrid zero dynamics \cite{Grizzle20103D}. Experiments are implemented on AMBER-3M planar robot with point feet, walking on a slippery treadmill. 
As a result, a conventional no-slip gait which walks successfully on a rough treadmill fails to walk on the slippery surface by losing its stability. On the other hand, a pre-planned gait which incorporates slippage walks successfully on the slippery surface, showing remarkable robustness with respect to treadmill speed as well as level of lubrication. Additionally, this gait even walks stably on the non-lubricated rough treadmill without slippage.

This manuscript is structured as follows: Section \ref{sec:model} consists of a detailed description of the hybrid walking dynamics. We stated all possible domains and edges associated with walking on slippery surfaces, where multiple switching guards are associated with one unique domain of dynamics. In Section \ref{sec:opt}, we briefly reveals the optimization algorithm and analysis the optimal trajectory's theoretical properties. Finally, Section \ref{sec:exp} and  \ref{sec:Conclusion} presented the experimental details of multiple successful AMBER-3M walking on slippery surfaces and provide analysis on the data.

\section{Hybrid dynamics}
\label{sec:model}

\begin{figure}[t]
\vspace{2mm}
	\begin{center}
		\includegraphics[scale=0.4]{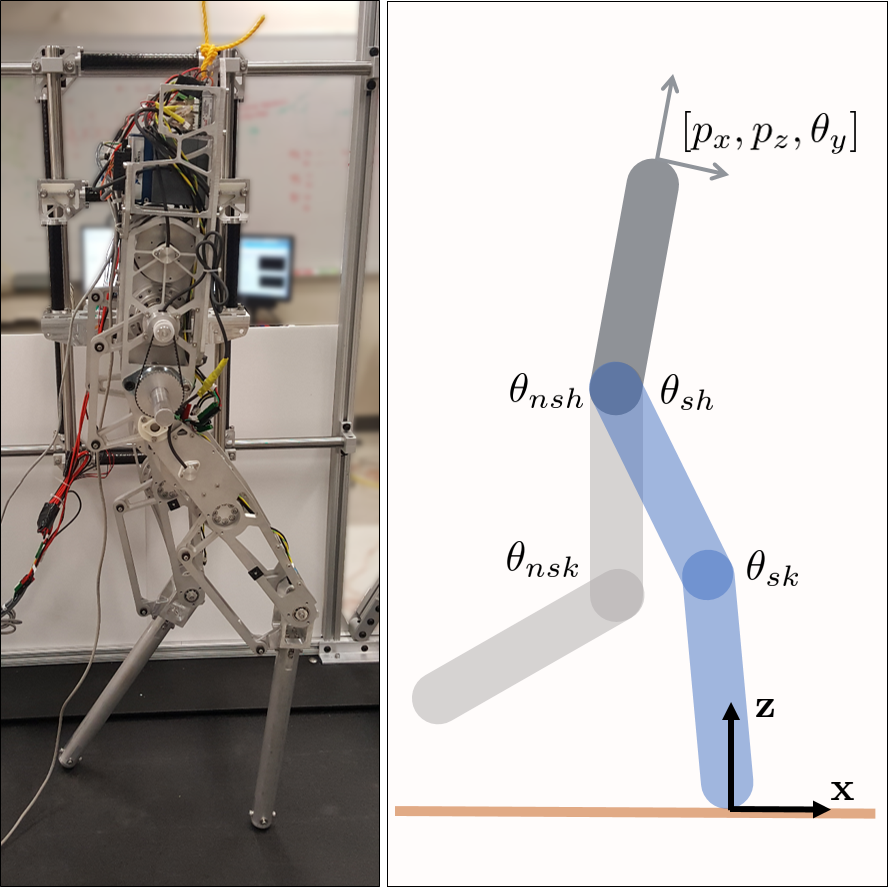}
		\caption{{\small On the Left: The AMBER-3M point foot version, constraint to a planar rail to walk in a 2D environment on a treadmill. On the right: the model's configuration coordinates, with 3 global coordinates and 4 local coordinates.}}
		\label{fig:robot}
	\end{center}
\vspace{-5mm}
\end{figure}

Walking on a slippery surface involves multiple continuous phases (or domains) that are related by discrete events; this naturally leads to a hybrid system model \cite{IntroHyb}. Therefore, this section presents the hybrid model corresponding to stick-slip walking gaits along with the stabilizing controllers.  

\begin{figure}[b]
	\vspace{-3mm}
	\centering
	\includegraphics[width=0.47\textwidth]{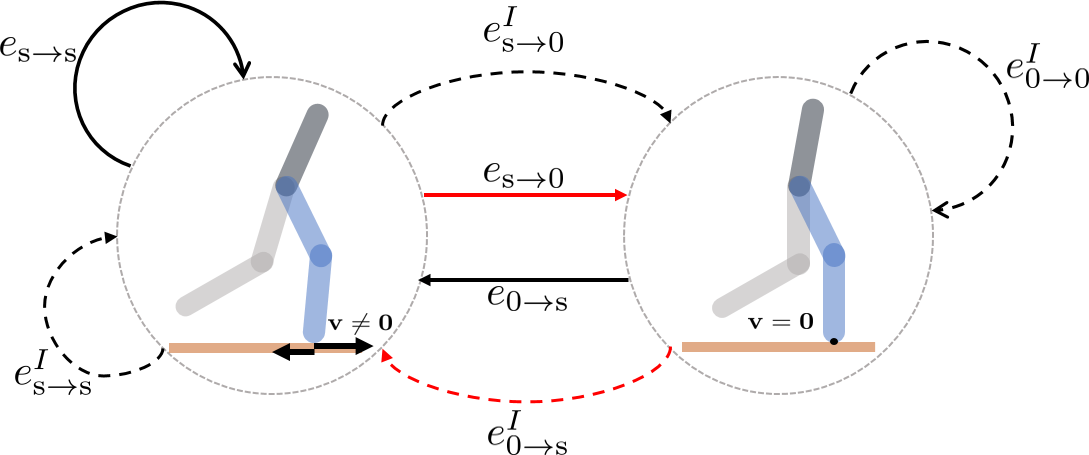}
	\caption{{\small The cyclic directed graph of the multi-domain hybrid system for walking on slippery surface. The solid lines are for transitions without non-stance foot impact events, and dash lines are for transitions with impact events.}}
	\label{fig:graph}
\end{figure}

\subsection{State and input space}
For the bipedal robot AMBER3-PF (PF is short for point foot, see \figref{fig:robot}), the configuration space is chosen as $q\in \mathcal{Q} \subseteq\R^n$, where $n$ is the number of unconstrained degrees of freedom (DOF), i.e. without considering contact constraints. Using the \textit{floating base convention} \cite{Grizzle20103D}, we have $q = (q_b, q_l)$, where $q_b$ are the global coordinates of the body fixed frame attached to the base linkage (torso), and $q_l$ are the local coordinates representing rotational joint angles and prismatic joint extensions. For planar walking on AMBER3-PF, it is chosen as $q_b = (p_x, p_y, \phi_y)$, where $p_x, p_y$ are the Cartesian positions of the torso and $\phi_y$ are the angle between the torso and world. The local coordinates are chosen as $q_l=(q_{sk}, q_{sh}, q_{nsh}, q_{nsk})$, each representing the stance knee, stance hip, non-stance hip and non-stance knee joint angle.

Further, the continuous-time state space $\mathcal{X}=T\mathcal{Q}\subseteq \R^{2n}$ has coordinates $x=(q^T,\dot q^T)^T$. Further, the control inputs $u\in\mathcal{U}\subseteq\R^m$ are for the actuator torques, with $m$ the total number of motors. For AMBER3-PF, we have $4$ motors on both knee and hip joints. This indicates under-actuated dynamics for AMBER-3M walking. 

\subsection{Hybrid System Model}
\label{sec:hybr-sys}
Due to the mixture of continuous and discrete dynamics, walking on a slippery surface is naturally modeled as a hybrid control system \cite{IntroHyb}. It is composed of two types of continuous domains: a \emph{sticky} walking, where the stance foot stays stationary at the contact point without moving while the other foot is swing in the air, and  \emph{slippery} walking, where the stance foot is slipping along $x$-axis on the ground. In summary, we define the walking dynamics on slippery surfaces as a \emph{tuple} \cite{Ames2013Humana}:
\begin{align}
	\label{eq:hybrid-control-system}
	\mathscr{HC} = (\Gamma, \mathcal{D}, \mathcal{U}, \mathcal{S}, \Delta, FG)
\end{align}
where,
\begin{itemize}
	\item $\Gamma = \{V, E\}$ is a directed graph with vertices $V=\{\mathrm{0,s} \}$, where $\mathrm{0}$ represents \textit{sticky} walking, and $\mathrm{s}$ represents slippage of the stance foot. The graph's edges are given by $E = \{e_{\mathrm{0\rightarrow s}},\; e_{	\mathrm{s\rightarrow 0}},\;  e_{\mathrm{s\rightarrow s}},\; e^I_{\mathrm{0\rightarrow 0}},\; e^I_{	\mathrm{0\rightarrow s}} ,\; e^I_{\mathrm{s\rightarrow 0}},\;\\ e^I_{	\mathrm{s\rightarrow s}} \}$. The superscript '$I$' denote transition via impact, whereas its absence denotes stick $\leftrightarrow$ slip transitions.
	\item $\mathcal{D} =\{ \mathcal{D}_\mathrm{0}, \mathcal{D}_\mathrm{s}\} \subseteq T\mathcal{Q}\times\mathcal{U}$ is a set of admissible domains of continuous dynamics,
	\item $\mathcal{U} =\{ \mathcal{U}_\mathrm{0}, \mathcal{U}_\mathrm{s}\}$ is a set of admissible controls,
	
	\item $\mathcal{S}=\{S[e]:$\hspace{0.1mm}$e \in E\}$ is a set of \textit{guards} referring to the switching surfaces between domains, which are associated with transitions represented by the directed edges in $E$.
	
	\item $\Delta = \{\Delta[e]:$\hspace{0.1mm}$ e\in E\}$ 
	is a set of smooth \emph{reset} \emph{maps} representing the discrete jump in states at each transition.

	\item $\emph{FG} = \{ (f_{\mathrm{0}},g_{\mathrm{0}}), (f_{\mathrm{s}},g_{\mathrm{s}})  \}$ is a set of  affine control systems $\dot x = f_v(x) + g_v(x)u$, defined on a domain $\Domain_{\vi}$.
\end{itemize}
The directed graph $\Gamma$ is depicted in the \figref{fig:graph}. 
It is very important to note, the previous research on \textit{multi-domain hybrid dynamics} such as \cite{zhaomulti, reher2016algorithmic,kolathaya2015composing} have an unique order of dynamics and events, this is due to the fact each domain is only associated with one unique event. But walking on slippery surfaces contains infinite possible sequence of motions. For example, the dynamics can transit from one domain to another through any of the events. The construction of individual elements of \eqref{eq:hybrid-control-system} will be explained in detail below. 

\subsection{Continuous-time dynamics for stick/slip domains}
A generalized representation of the stick-slip walking dynamics can be modeled using the \textit{constrained Lagrangian dynamics} \cite{Murray1994mathematical,gamus2015dynamic}. A kinematic constraint of zero normal displacement of the stance foot reads as $z_s(q)=0$. An additional no-slip constraint in tangential direction occurs only in the {\it stick} domain, and is given by $x_s(q)=x_0$. 
For a particular continuous domain $(q, \dot q)\in\mathcal{D}_v$, the dynamics is formulated as
\begin{align}
\label{eq:eom-general}
	D(q) \ddot q + H(q,\dot q) &= B u + J_{x}^T(q) \lambda_{x} + + J_{z}^T(q) \lambda_{z}
\end{align}
where, $D(q)\in\R^{n\times n}$ is the inertia matrix, $ H(q,\dot q)\in\R^n$ contains the Coriolis, gravity forces, and $B(q)$ is the actuation matrix, all of which are given by the physical parameters of the robot and remain the same across all continuous domains. In addition, the Jacobian matrices (constraint gradient vectors) in \eqref{eq:eom-general} are defined as $J_x(q)=\dfrac{d x_s(q)}{dq}$ and $J_z(q)=\dfrac{d z_s(q)}{dq}$, and $\lambda_x, \lambda_z$ are the tangential and normal forces enforcing the contact constraints. In the domain of sticking contact, expressions for the contact forces can be obtained by augmenting the second time-derivative of the constraints:
\begin{align}
\dot J(q,\dot q)\dot q + J(q) \ddot q =0 \mbox{ , where } J(q)= \left( \begin{array}{l} J_x(q) \\ J_z(q) \end{array} \right).
\label{eq:Jdd}
\end{align}
Eliminating $\ddot q$ from \eqref{eq:eom-general} and substituting into \eqref{eq:Jdd}, one can solve for the constraint forces under sticking contact (cf. \cite{Murray1994mathematical,gamus2015dynamic}):
\begin{align}
\left( \begin{array}{l} \lambda_x^0 \\ \lambda_z^0 \end{array} \right) = \left(J D^{-1} J^T \right)^{-1}\left(JD^{-1}(H-Bu) - \dot J \dot q \right),
\label{eq:lambda_stick}
\end{align}
where the dependencies on $q,\dot q,u$ in \eqref{eq:lambda_stick}  are suppressed for brevity. The forces must satisfy Coulomb's inequality of dry friction:
\begin{align}
|\lambda_x^0(q,\dot q,u)| \leq \mu  \lambda_z^0(q,\dot q,u),
\label{eq:ineq_fric}
\end{align}
where $\mu$ is the {\it coefficient of friction}. When the friction is too low, slippage of the stance foot in tangential direction begins to evolve, $\dot x_{s} = J_x \dot q \neq 0$. In this case, the equation of motion \eqref{eq:eom-general} still holds while the tangential constraint in \eqref{eq:Jdd} is no longer valid. Instead, the following two equations should be augmented with \eqref{eq:eom-general}:
\begin{align}
	\dot J_z(q,\dot q)\dot q + J_z(q) \ddot q =0 \\[5pt]
	\lambda_x  = -\mathrm{sgn}(\dot x_s) \mu \lambda_z
	\label{eq:Jzdd}
\end{align}
The tangential force during slippage reaches its maximal magnitude while opposing the slip direction. (Note that we do not distinguish here between static and dynamic friction coefficients for simplicity). Combining \eqref{eq:eom-general} and \eqref{eq:Jzdd} to obtain expressions for the constraint forces during slippage (\cite{gamus2015dynamic}):
\begin{align}
\lambda_z^s(q,\dot q,u) =& \left(J_z D^{-1} (J_z  -\mathrm{sgn}(\dot x_s) \mu J_x)^T \right)^{-1} \notag \\
                         & \left(JD^{-1}(H-Bu) - \dot J_z \dot q \right)\\
\lambda_x^s(q,\dot q,u) =&-\mathrm{sgn}(\dot x_s) \mu \lambda_z.
\label{eq:lambda_slip}
\end{align}

Inequality constraints for slippage are $\lambda_z \geq 0$ and $\dot x_s \neq 0$. Finally, in both domains the non-stance foot must stay above the ground, $z_{ns}(q) \geq 0$. We can now introduce both stick and slip domains and their definition:
\begin{align*}
	\mathcal{D}_0=\{ & (x,u)\in T\mathcal{Q}\times \mathcal{U} \ | \;\;
						z_s=\dot z_s=\dot x_s =0, \notag\\ 
					& \hspace{2.9cm} z_{ns} \geq 0, \; | \lambda_x^0| \geq \mu \lambda_z^0\} \tag{*} \\[3pt]
	\mathcal{D}_s=\{ & (x,u)\in T\mathcal{Q}\times \mathcal{U} \ | \;\;
		   			   z_s=\dot z_s=0, \notag\\
					& \hspace{2.9cm} \dot x_s \neq 0,\; z_{ns} \geq 0, \;  \lambda_z^s \geq 0 \} \tag{+}
\end{align*}
Finally, for a particular domain $v\in\{0,\mathrm{s}\}$, we can convert the dynamics \eqref{eq:eom-general} and constraint forces in \eqref{eq:lambda_stick} or \eqref{eq:lambda_slip} into an affine control system in state space \cite{ames-hi} as:
\begin{align}
	\label{eq:olsystem}
	\dot x = f_v(x) + g_v(x)u\ \ \ \ \ \forall x\in\mathcal{D}_v
\end{align}

\subsection{Discrete dynamics}
The reset maps associated with non-impacting transitions $\Delta[e_{\mathrm{0\rightarrow s}}],\Delta[e_{\mathrm{s\rightarrow 0}}],\Delta[e_{	\mathrm{s\rightarrow s}}]$ are simply an identity matrix: $x^+ = x^-$, where $x^-$,$x^+$ are the pre-event and post-event states. This means that the transition is smooth in state space. In the case of collision of the non-stance foot, the transition involves impact which induces an instantaneous velocity jump $\dot q^+ = \Delta[e] \dot q^-$. The impulse-momentum balance reads as follows (cf. \cite{Grizzle2014Models,gamus2015dynamic})
%
\begin{align}
	\label{eq:impact1}
	D(q_c) (\dot q^+ -\dot q^-) &= J(q_c)\Lambda = J_x(q_c)^T \Lambda_x + J_z(q_c)^T \Lambda_z
\end{align}
where $q_c$ is the robot's configuration at collision and $\Lambda=(\Lambda_x,\Lambda_z)^T$ are tangential and normal impulses at the colliding foot. (Note that one has to interchange the stance and non-stance variables right before impact, so that $J_x,J_z$ are associated with velocities of the colliding foot.) The commonly used model is that of perfectly inelastic impact. Assuming zero tangential and normal contact velocities at the post-impact state gives $J(q_c)\dot q^+=0$. Combining this with \eqref{eq:impact1}, one obtains the contact impulse and post-impact velocity as:
\begin{align*}
	\Lambda^0=\left( \begin{array}{l} \Lambda_x^0 \\ \Lambda_z^0 \end{array} \right)=-(JD^{-1}J^T)^{-1}J \dot q^- \\[3pt]
	\dot q^+=\left(I - D^{-1}J^T (JD^{-1}J^T)^{-1}J \right)\dot q^-
\end{align*}
where $I$ is the identity matrix and $D,J$ are evaluated at $q=q_c$. This is the sticking impact law, associated with reset maps $\Delta[e]$ for transition edges $e^I_{\mathrm{0\rightarrow 0}}, e^I_{\mathrm{s\rightarrow 0}}$. This solution holds only if the impulses satisfy the frictional inequality $|\Lambda_x^0| \leq \mu \Lambda_z^0$. Otherwise, a slipping impact occurs where $J_z \dot q^+ =0$ while $J_x \dot q^+ \neq 0$. The impulses are thus related as $\Lambda_x=-\mathrm{sgn}(J_x(q_c) \dot q^+) \mu \Lambda_z$. Combining this with \eqref{eq:impact1}, one obtains
\begin{align*}
	\Lambda_z^s &=-(J_z D^{-1}\tilde J^T)^{-1}J_z \dot q^- \\[3pt]
	\dot q^+    &=\left(I - D^{-1}\tilde J^T (J_z D^{-1}\tilde J^T)^{-1}J_z \right)\dot q^- 
\end{align*}
where $\tilde J = J_z -\mathrm{sgn}(J_x(q_c) \dot q^+)  \mu J_x$. This slipping impact law is associated with reset maps $\Delta[e]$ for transition edges $e^I_{\mathrm{0\rightarrow s}}, e^I_{\mathrm{s\rightarrow s}}$.

\subsection{Guards}
We now define the guards, which are switching surfaces or conditions for transition between domains. The first guards are associated with the smooth transitions between sticking and slipping domains:
\begin{align*}
\mathcal{S}[e_{	\mathrm{0\rightarrow s}}] &= \{(x,u)\in \mathcal{D}_\mathrm{0} \ |\ |\lambda_x^0| = \mu \lambda_z^0  \} \\[3pt]
\mathcal{S}[e_{	\mathrm{s\rightarrow 0}}] &= \{(x,u)\in \mathcal{D}_\mathrm{s} \ |\ \dot x_s=0,\; |\lambda_x^0| \leq \mu \lambda_z^0  \} \\[3pt]
\mathcal{S}[e_{	\mathrm{s\rightarrow s}}] &= \{(x,u)\in \mathcal{D}_\mathrm{s} \ |\ \dot x_s=0,\; |\lambda_x^0| > \mu \lambda_z^0  \}
\end{align*}
Note that the last transition above associated with $e_{	\mathrm{s\rightarrow s}}$ is reversal of slip direction (cf. \cite{gamus2015dynamic}). The guards corresponding to transitions that involve sticking or slipping impacts are defined as:
\begin{align*}
\mathcal{S}[e^I_{\mathrm{0\rightarrow 0}}] = \{(x,u)\in \mathcal{D}_\mathrm{0} \ |\; z_{ns}=0, \dot z_{ns}<0 \mbox{ and } |\Lambda_x^0| \leq \mu \Lambda_z^0 \} \\[3pt]
\mathcal{S}[e^I_{\mathrm{s\rightarrow 0}}] = \{(x,u)\in \mathcal{D}_\mathrm{s} \ |\; z_{ns}=0, \dot z_{ns}<0 \mbox{ and } |\Lambda_x^0| \leq \mu \Lambda_z^0 \} \\[3pt]
\mathcal{S}[e^I_{\mathrm{0\rightarrow s}}] = \{(x,u)\in \mathcal{D}_\mathrm{0} \ |\; z_{ns}=0, \dot z_{ns}<0 \mbox{ and } |\Lambda_x^0| > \mu \Lambda_z^0 \} \\[3pt]
\mathcal{S}[e^I_{\mathrm{s\rightarrow s}}] = \{(x,u)\in \mathcal{D}_\mathrm{s} \ |\; z_{ns}=0, \dot z_{ns}<0 \mbox{ and } |\Lambda_x^0| > \mu \Lambda_z^0 \} 
\end{align*}
These guards represent the conditions for sticking or slipping impacts as described above. Note that the overall non-smooth frictional dynamics may have special degenerate cases where the solution is inconsistent, indeterminate, or singular. These rare cases are know as {\it Painlev\'{e} paradox} \cite{champneys2016painleve, or2014painleve}, and lie beyond the scope of this work.

\subsection{Feedback Controllers}
To stably control a continuous domain $\mathcal{D}_v$ with $v\in\{0,\mathrm{p}\}$, we used virtual constraint based walking controllers \cite{Ames2012First}. In this controller, we first define a set of virtual holonomic constraints:
\begin{align}
	y(q) = y^d_{\alpha_v}(\tau(q)) - y^a(q)
\end{align}
where $y^d_\alpha(\tau(q))$ is the desired trajectory for the chosen features defined by $y^a(q)$; in this case, simply the four actuated joints: $y^a(q):=(\theta_{sk}, \theta_{sh}, \theta_{nsh}, \theta_{nsk})^T$. We use a monotonically increasing \textit{phase variable} $\tau(q)$ to parameterize the trajectory. The trajectory is described by a set of static parameters $\alpha_v\in\R^{m\times 5}$ for each domain. Then we applied a standard input-output feedback linearization to drive $y\rightarrow 0$ exponentially, more details can be found in \cite{Ames2012First}. The key idea of this methodology is to remove time dependency and control the dynamics to evolve naturally along the zero dynamics, i.e., the under-actuated dynamics.
\section{An optimization formulation}
\label{sec:opt}

To generate a slippery walking gait, we formulate this control problem as an implicit trajectory optimization problem. In particular, we used a direct collocation method to factorize it into a regular nonlinear programming (NLP). A deep review of collocation methods can be found in \cite{kellyDC}. In essence, this method numerically solves nonlinear dynamics by minimizing the difference between the approximate and exact solutions at collocation points. Formally, 
\begin{align}
\label{eq:opteqs}
	\min_{\alpha, x_i, \dot x_i u_i}  &\hspace{3mm}  u_i^T u_i \hspace{1cm}i\in\{1,2,...2M+1\}\\
	\mathrm{s.t.} 		&\hspace{3mm}  \textbf{C1}.\ \mathrm{closed\ loop\ dynamics} \notag \\
				  		&\hspace{3mm}  \textbf{C2}.\ \mathrm{hybrid\ periodic} \notag\\
				  		&\hspace{3mm}  \textbf{C3}.\ \mathrm{physical\ limitations} \notag\\
						&\hspace{3mm}  \textbf{C4}.\  \mathrm{slipping\ feasibility} \notag
\end{align}
with $M$ the total number of collocation points, and the target is to minimize torque inputs. For our problem, FROST (\textit{Fast Robot Optimization and Simulation Toolkit}) \cite{hereid2018dynamic} was employed for its robustness in solving closed-loop trajectory optimization problems.

\subsection{Constraints}
In the FROST formulation, the optimization picks an optimal \textit{state and input trajectory} such that they satisfy the closed-loop dynamics at each node $i$ by using hermite-simpson method \cite{Hereid2016}. This is represented by the \textit{closed loop dynamics} constraints \textbf{C1}:
\begin{align}
	\dot x_i &= f_v(x_i) + g_v(x_i) u_i \label{eq:idyn}\\
	\ddot y_i &= -2\epsilon \dot y_i -\epsilon^2 y_i \label{eq:iydyn}
\end{align} 
for $x_i\in\mathcal{D}_v$. Note that equality constraint \eqref{eq:iydyn} eliminates the need to invert dynamics \eqref{eq:idyn} for the state dependent controller $u(x)$. We also enforced periodic constraint \textbf{C2} so that the states at the edges of each domain are connected. For real world implementation, we also considered physical constraints \textbf{C2}, such as limiting torques less than $40$Nm, joint velocity less than $4$rad/s and preventing certain joint to hyperextend. Put simple, \textbf{C1-C3} solves the given nonlinear dynamics in an optimal way. 
But to yield a slipping gait, we additionally includes feasibility constraints \textbf{C4} from definitions in (*) (+). In our formulation, we pre-specified a specific ordered sequence of transitions, indicated by the red line in \figref{fig:graph}.
Additionally, since a smoother state trajectory is preferred for experiment robustness, we further forced the static parameters to be the same across all domains. It is worthwhile to mention, this constraint is feasible if and only if the transition between domains within one step does not involve any jump in states. This yielded a uniform trajectory for the multi-domain walking dynamics.

\normalsize
\subsection{Optimal gaits}
\begin{figure}[t]
\vspace{1.8mm}
	\centering
	\includegraphics[width=0.44\textwidth]{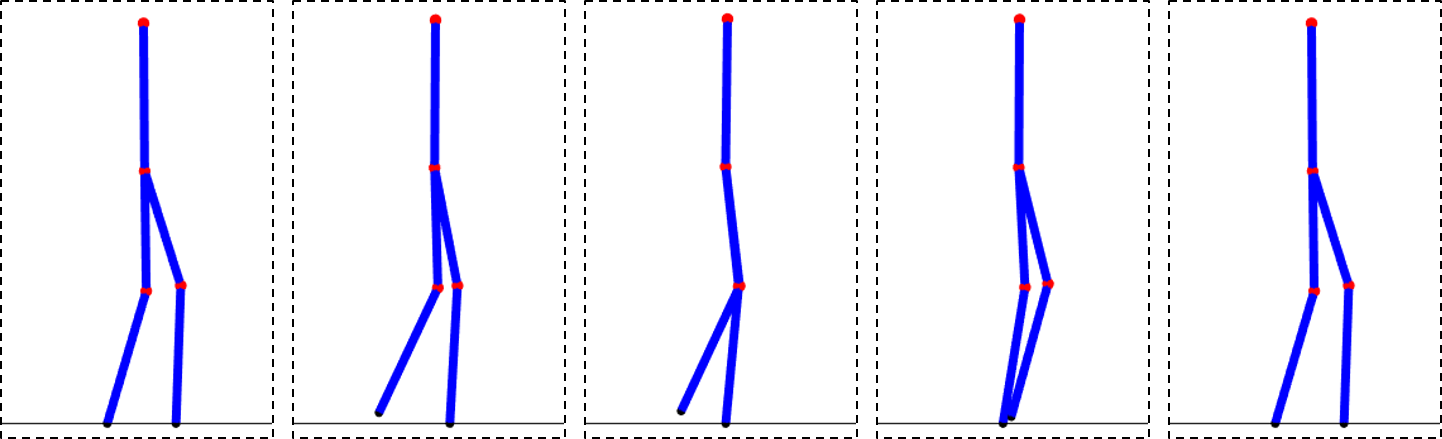}
	\caption{The slippery walking gait from optimization.}
	\label{fig:slipperytiles}
\vspace{-4mm}
\end{figure}
Solving the optimization problem \eqref{eq:opteqs}, we obtained a two-domain slippery walking gait with slippage on the stance foot $3$cm, shown with snapshots in \figref{fig:slipperytiles}. The $MCOT^+$ from optimization is given as $0.001$. 
The \textit{positive only mechanical cost of transport} is calculated using 
\begin{align}
	MCOT^+ = \dfrac{  \bar P}{mgv}
	\label{eq:mcot}
\end{align}
with $m$ the total mass, $g$ the gravitational acceleration, $v$ the average walking speed, and $\bar P^+$ is the mean value of $P^+= \{ P_i^+ \}_{i=1}^N$ with $i\in\{1,2,3...N\}$ and $N$ is the total number of sample points. The positive only power at sample time $t_i$ is computed by 
$P^+_i = \sum_{k=1}^4 \max\big(u_i(k) \cdot \dot q_i(k),0\big), $
with $u_i\in\R^4$ and $q_i\in\R^4$ torques and velocities of the actuated joints at time $t_i$.  

For a fair comparison against \textit{sticky walking}, we simulate the slippery gait based controller in a sticky environment, i.e., the ground has a much higher friction coefficient so that no slipping can happen. 
After $20\sim 30$ steps, the walking converged into a new stable patten. 
with $MCOT^+$ being $0.0024$. 
which is $140\%$ less energy efficient than walking on a slippery surface. 
Further at its steady state, the non-stance foot's velocities changed from $(0.563, -0.359)$m/s to $(0,0)$m/s through the sticky impact. The body kinetic energy changing from $6.87$J to $5.84$J. 
However, the original optimal slippery gait has a non-stance foot impact velocity changing from $(0.371, -0.237)$m/s to $(0.251, 0)$m/s, and kinetic energy changing from $3.00$J to $2.63$J. 
This aligns with the theories on energy efficiency in \cite{gamus2015dynamic}.

\section{Realization}
\label{sec:exp}

AMBER-3M is a modularized testbed to study planar bipedal locomotions. Its robustness and durability was validated in multiple experiments \cite{ambrose2017toward, tabuada2017data}. In this paper, we particularly studied the slippery walking behavior on the point foot version (with total mass $21.6$kg). As detailed in \cite{ambrose2017toward}, the planar walking is achieved by constraining the robot on a planar rail structure and walking on a treadmill (\figref{fig:robot}). Further experimental details will be presented in this section.

\subsection{Experimental controller}
\begin{figure}[!t]
	\vspace{2mm}
	\begin{center}
		\includegraphics[width=0.43\textwidth]{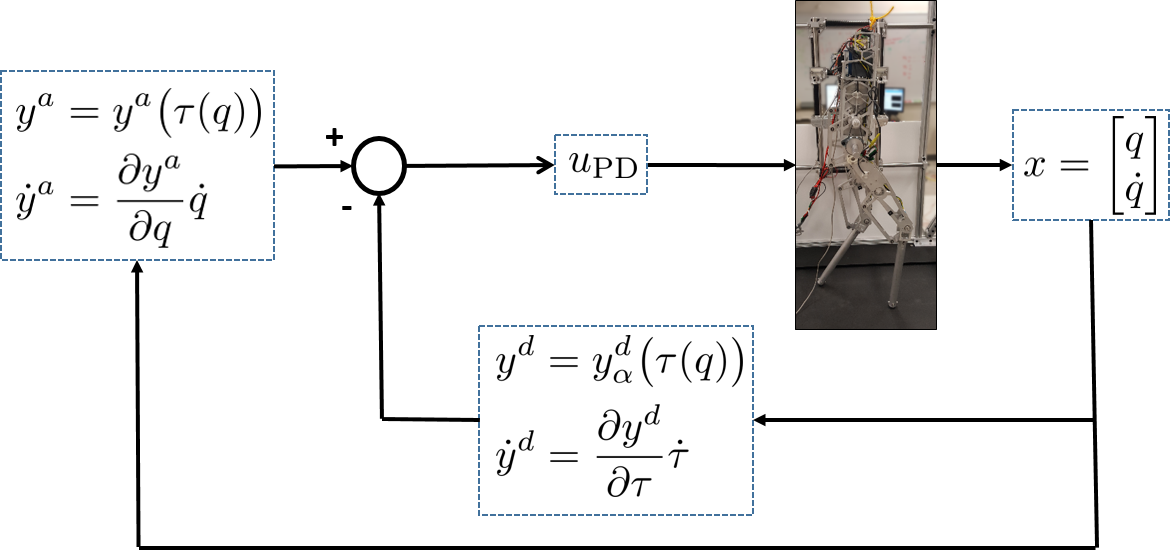}
		\caption{{\small Block diagram for the controller implemented on AMBER3-PM for the walking gaits considered in this paper.}}
		\label{fig:expControl}
	\end{center}
\vspace{-6mm}
\end{figure}

The control structure used in experiments for walking on a slippery surface is shown in \figref{fig:expControl}. In this block diagram, we first measures each joint's position and velocity, including the global orientation of the torso. Then the \textit{phase variable} $\tau(q)$ can be obtained. Note that due to the slippage of the stance foot, we used the \textit{linearized-relative hip position} to calculate the phase variable:
\begin{align*}
	\tau(q) = \dfrac{p(q) - p_0}{p_1 - p_0}
\end{align*}
with $p(q) = \delta(hip_\mathrm{x} - sf_\mathrm{x})$ and $hip_\mathrm{x}$ the hip joint's Cartesian position calculated from the global coordinates origin, $p_0$ and $p_1$ are its initial and final value. This way, the phase variable is independent of the noisy measurement of the foot slippage. Next, the desired outputs are calculated based on the optimal trajectory parameters $\alpha$. Together with the actual outputs $y^a(q)$, on the robot we use a PD controller:
\begin{align*}
	u_\mathrm{PD} = k_p(y_a-y_d) + k_d(\dot y_a - \dot  y_d)
\end{align*} 
instead of the feedback linearization controller assumed in the optimization algorithm. This implementation difference between theory and experiment has been justified for improved robustness in \cite{kolathaya2017parameter}.

\normalsize
\subsection{Experiments}
\begin{figure*}[t]
	\vspace{2mm}
	\begin{center}
		\includegraphics[width=0.9\textwidth]{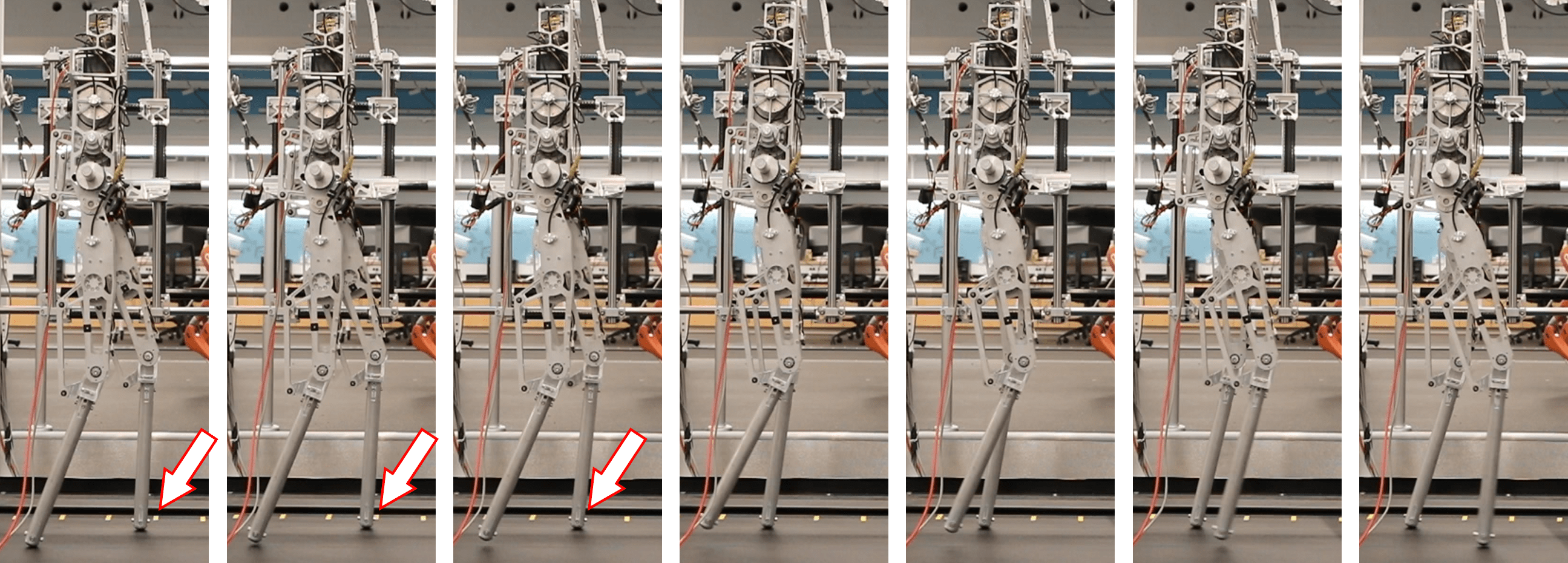}
		\caption{{\small Snapshots of one slippery walking step from \textit{Experiment 3}. In the first three pictures, the left foot (stance foot) is slipping smoothly on the lubricated treadmill.}}
		\label{fig:snapshot}
	\end{center}
	\vspace{-5mm}
\end{figure*}

\begin{figure*}[t]
\vspace{-2mm}
	\centering
		\includegraphics[width=0.215\textwidth]{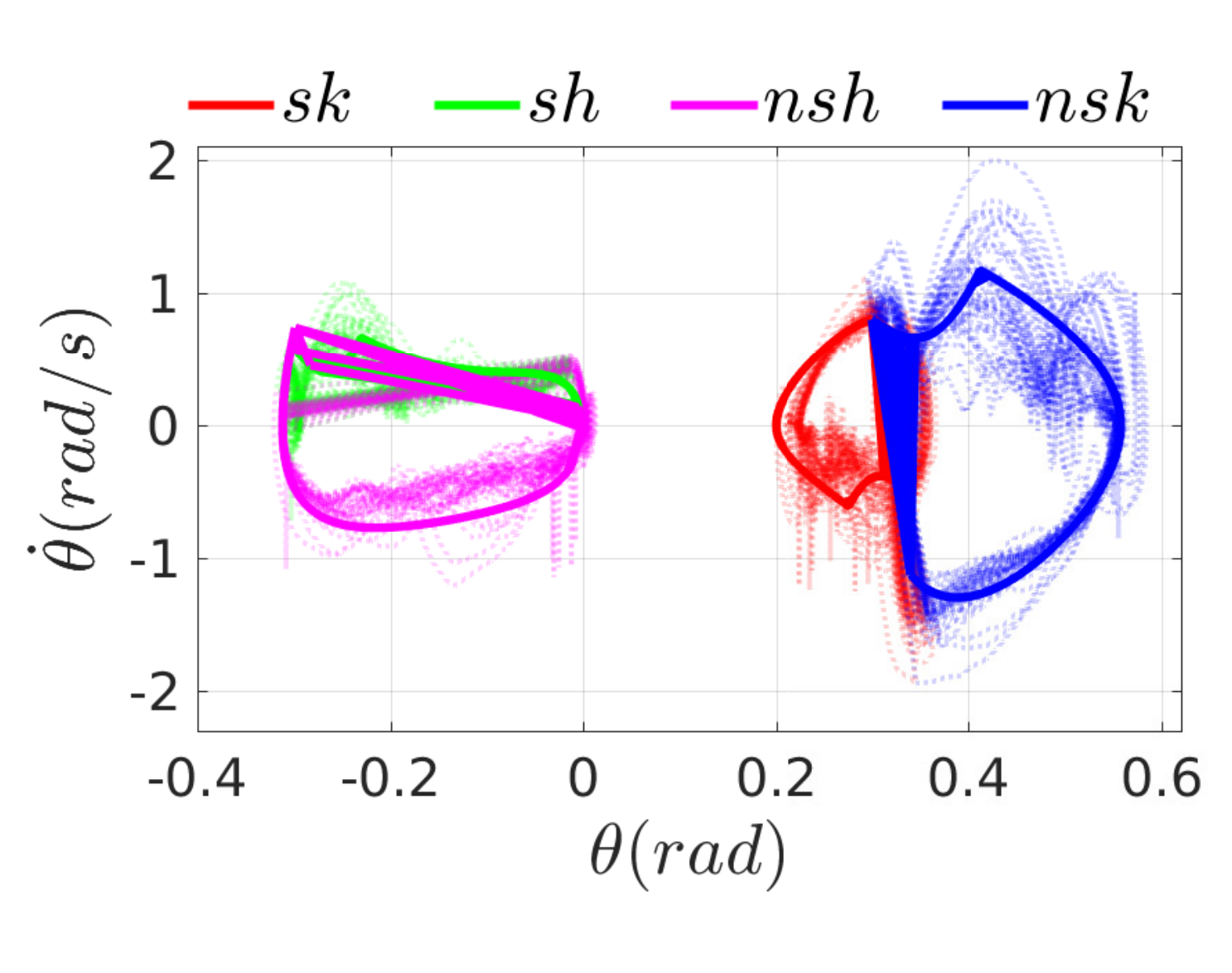}
		\includegraphics[width=0.215\textwidth]{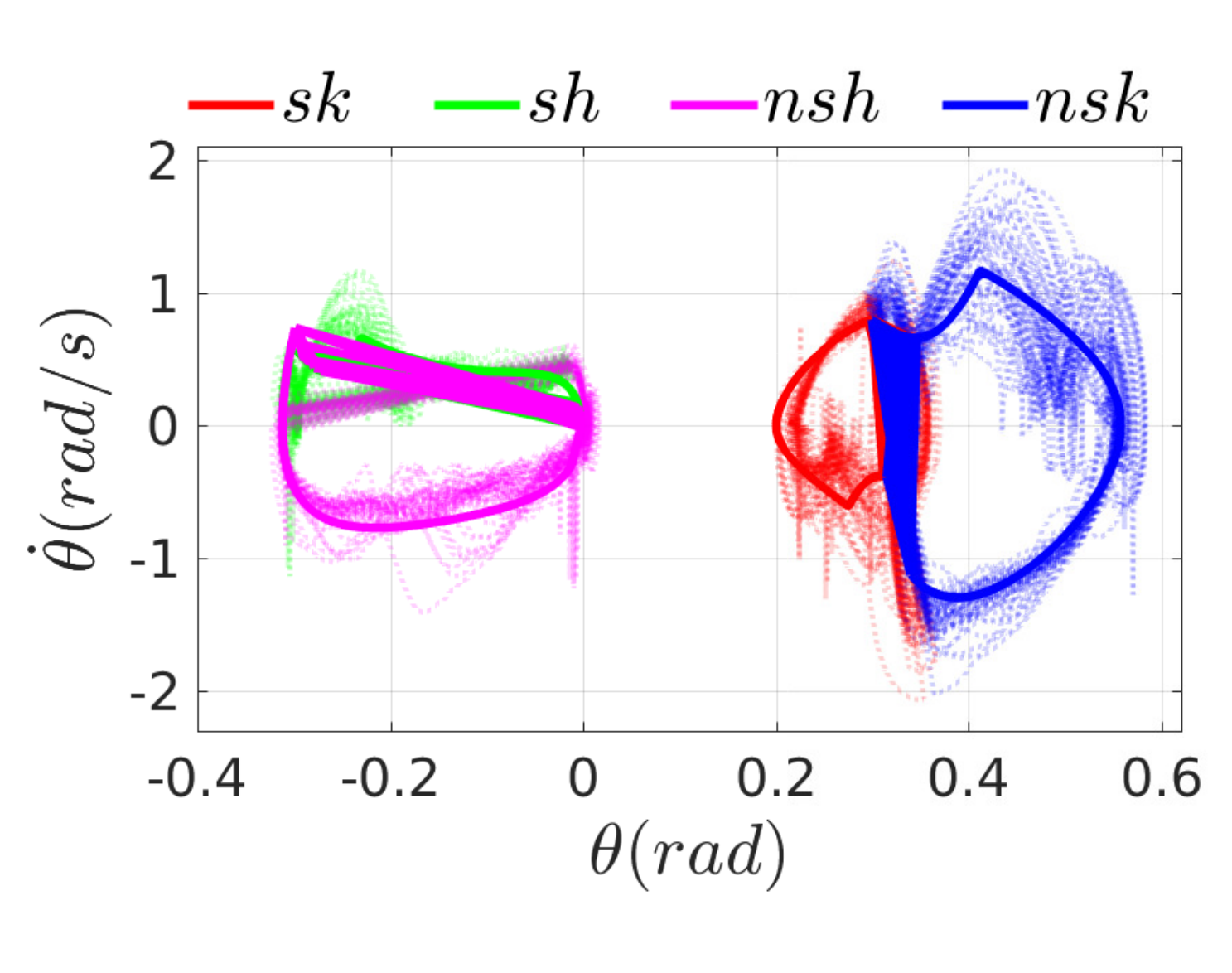}
		\includegraphics[width=0.23\textwidth]{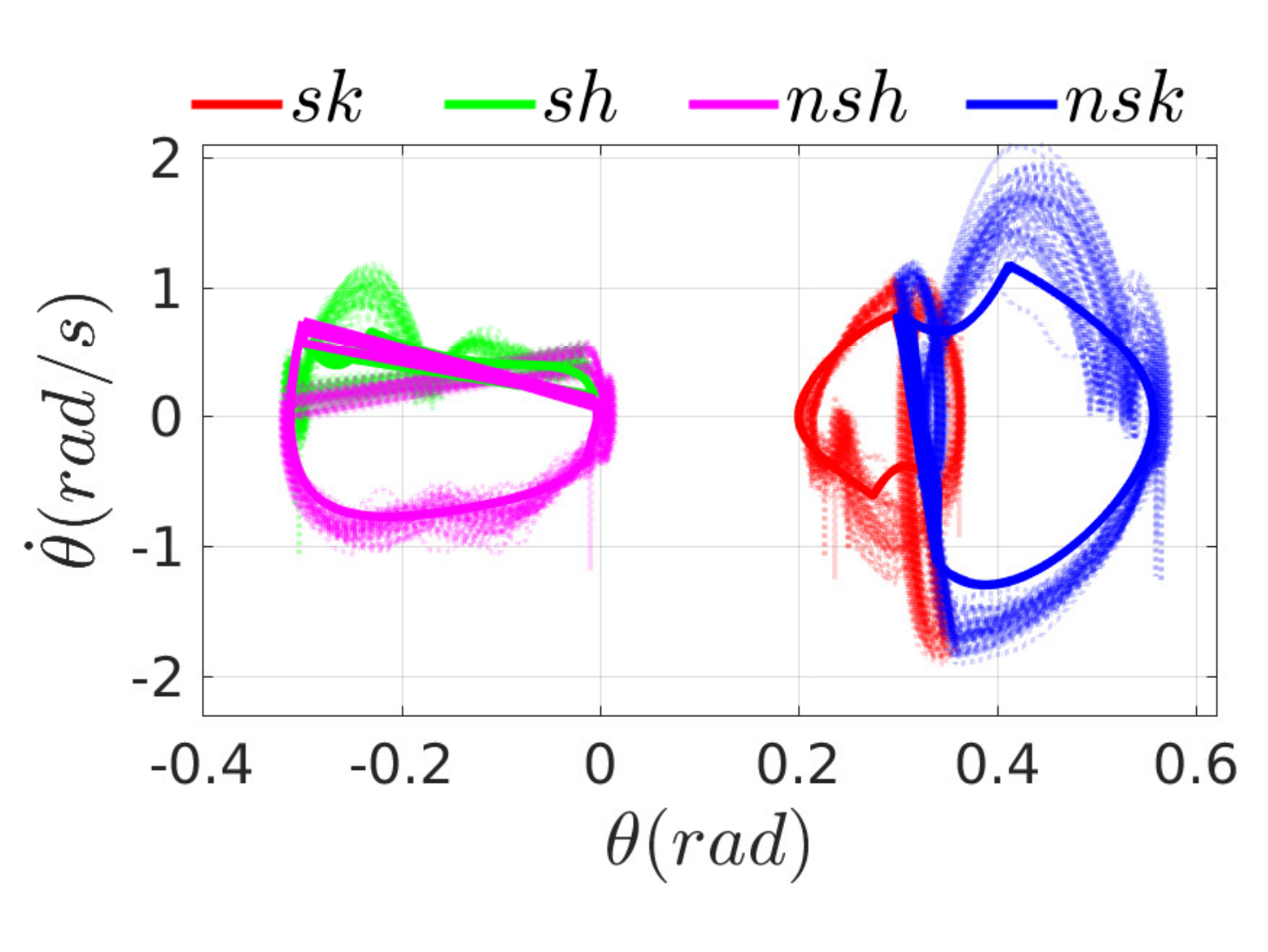}
		\includegraphics[width=0.23\textwidth]{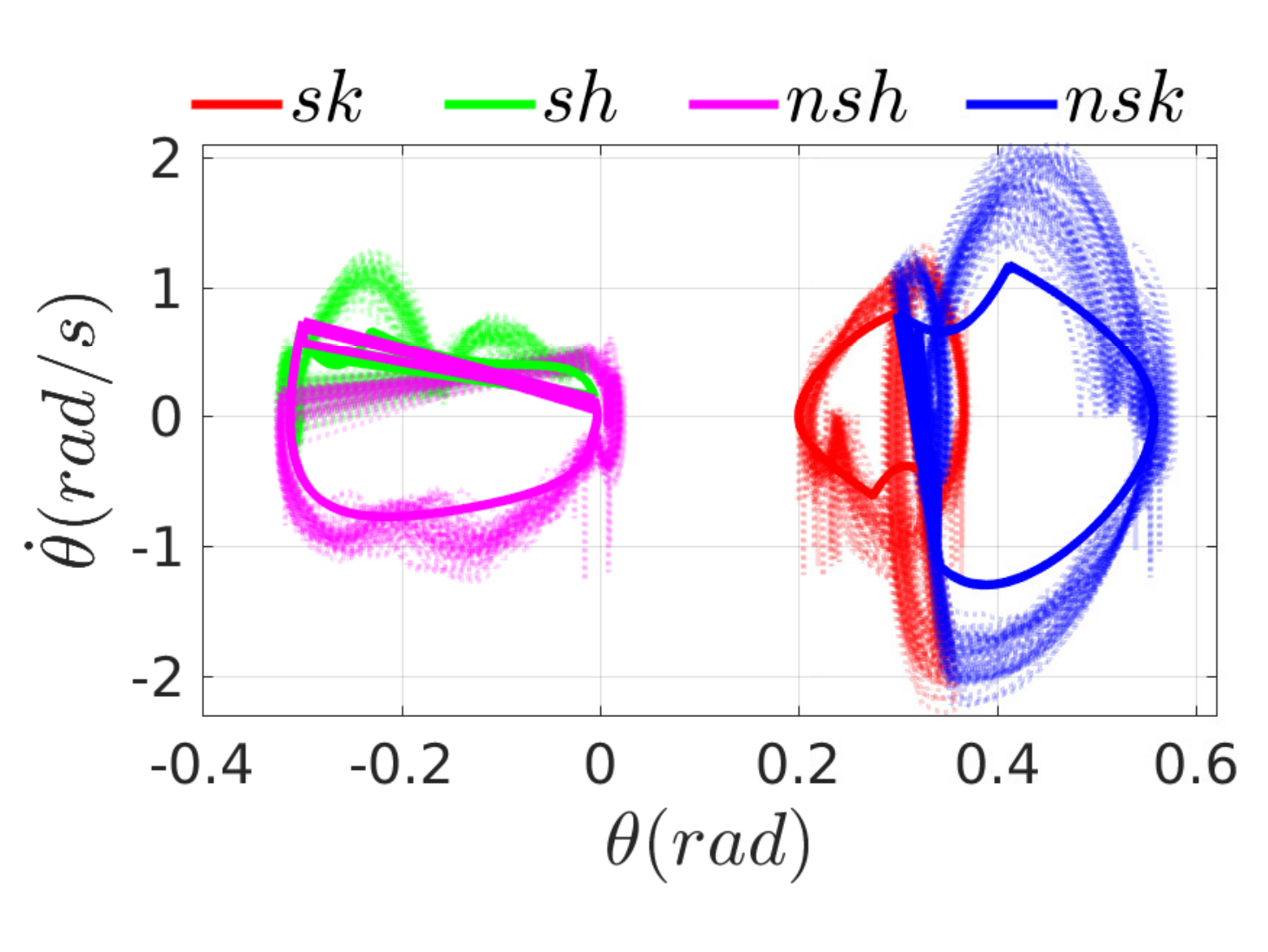}
	\vspace{-3mm}
	\caption{{\small Phase portrait of $50$ seconds' experimental data from \textit{Experiment} $3$, with walking speed (from left to right): $0.26$m/s, $0.3$m/s, $0.38$m/s, $0.42$m/s. Solid lines are for the desired values and dash lines are for the actual measurements.}}
	\label{fig:pp}
	\vspace{-5mm}
\end{figure*}

To begin with, we placed a ``demonstration'' walking gait that was designed for a sticky surface on the slippery surface covered by lubricant. Its robustness has been challenged by countless visitors with failure (i.e., lose of stability) almost never occurring. However, a few drops of lubricant easily disabled its walking capability (see \cite{fail}). To clarify, we consider falling down and hitting the mechanical limits of the testbed both as failures. Later, we conducted four different experimental setups. For \textit{experiment} $1,2$ and $3$, we increased the amount of lubricant on the treadmill to induce different slippery walking behaviors, and completely removed the lubricant for \textit{experiment} $0$. For each fixed environmental setup, we manually increase the treadmill speed to trigger different walking speeds on the slippery surface. We logged $50$ seconds' data (sampling period $3$ms) for each experiment to calculate the energy economy. \figref{fig:pp} shows the phase portrait in for \textit{experiment} $3$ which has the most slippery surface. The result is AMBER-3M is capable of walking stably on different slipping conditions including on a sticky surface, proving its robustness and adaptability to uncertainties between simulation and experiments. See \cite{walk} for the robust walking on slippery surfaces.

\subsection{Energy economy}

\begin{figure}[!t]
	\vspace{2mm}
	\centering
		\includegraphics[width=0.35\textwidth]{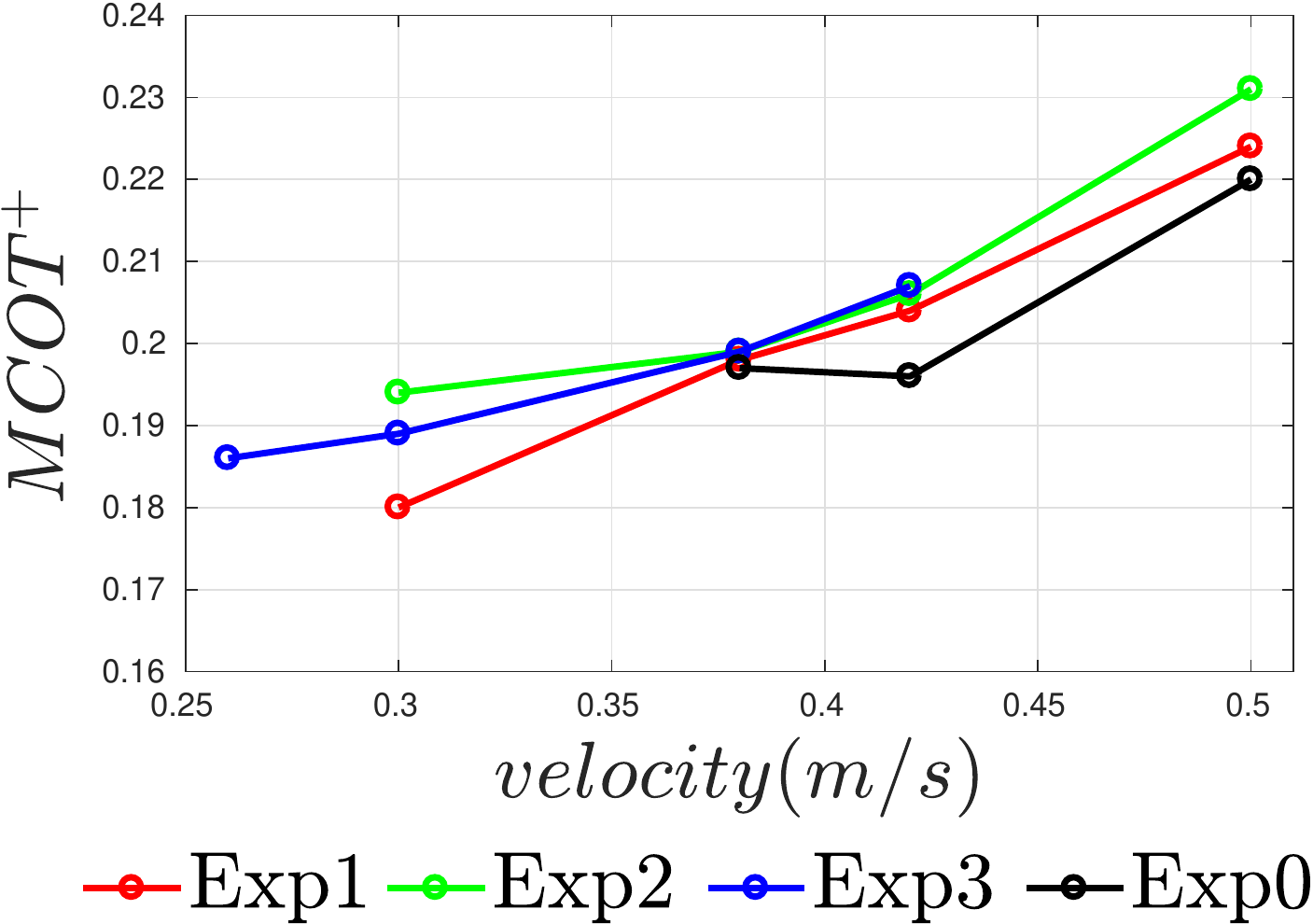}
		\caption{{\small$MCOT^+$ of all experiments. Those not included for certain speeds are failed experiments.}}
	\label{fig:cot}
	\vspace{-5mm}
\end{figure}

A previous research \cite{ambrose2017toward} on AMBER-3M with a circular boom  has a benchmark on the energy economy of walking controllers. In this research due to the slippage of stance foot, it became too noisy to measure the absolute movement of the \textit{Center of Mass}. Hence we used the measure \eqref{eq:mcot} for $MCOT^+$. See \figref{fig:cot} for the energy results. Note that we only provide \textit{positive only power} because AMBER-3M's hardware cannot do power-regeneration of the negative work.

While the energy efficiency \figref{fig:cot} seems better than \cite{ambrose2017toward}, our measure shows experiment energy efficiency is $\sim10$ times worse than simulation, and the efficiency on difference surfaces does not vary as much as simulation data. This is not only caused by different external environment such as inconsistency of the lubricated treadmill and real world uncertainties, 
but we posit the dominance of the $MCOT^+$ by nominal energy usage of the robot.  That is, due to the order of magnitude difference the simulation and experimental $MCOT^+$, the comparatively small fluctuations in the MCOT between different walking cannot be observed with the current experimental setup.  Therefore, it is necessary to study differences in the $MCOT^+$ between slipping and nominal gaits wherein changes in energy usage can be isolated from nominal energy usage and the effects of the environment on the cost of transport. This is the subject of future research. 

\section{Concluding Remarks}
\label{sec:Conclusion}
\normalsize
In this paper, we formally defined dynamical walking on slippery surfaces from a hybrid system perspective. This definition made it possible to formally decomposition this problem into a traditional trajectory optimization problem and, for the first time, we are able to demonstrate dynamically stable walking on slippery surfaces experimentally; this walking showed satisfying robustness and agility. Future work includes studying different ordered sequences of domains and more a more comprehensive study on  energy consumption for bipedal robots walking on slippery surfaces.

\bibliographystyle{IEEEtran}
\bibliography{citation}

\end{document}